\documentclass{article} 
\usepackage{hyperref}
\usepackage{url}
\usepackage{array}
\usepackage{amssymb,amsmath}
\usepackage{amsthm}
\usepackage{makecell}
\usepackage{mathtools}
\usepackage{longtable}
\usepackage{caption} 

\DeclarePairedDelimiter{\ceil}{\lceil}{\rceil}
\DeclarePairedDelimiter{\floor}{\lfloor}{\rfloor}

\usepackage{algorithm}
\usepackage{algorithmic}
\usepackage{graphicx}
\usepackage{csvsimple}
\usepackage[group-separator={,},
				group-minimum-digits=4,
				table-format = 7]{siunitx}

\author{
Mingyuan Wang, Adrian Barbu\\
Department of Statistics \\
Florida State University\\
}

\title{Online Feature Screening for Data Streams with Concept Drift}

\graphicspath{{pics/}}
\DeclareGraphicsExtensions{.jpg,.pdf,.mps,.png}

\def\RR{\mathbb R}

\newcommand{\bo}{\mathbf{o}}

\newcommand{\bx}{\mathbf{x}}

\newcommand{\by}{\mathbf{y}}

\newcommand{\bw}{\mathbf{w}}

\newcommand{\bbeta}{{\boldsymbol{\beta}}}

\newcommand{\be}{{\boldsymbol{\epsilon}}}

\begin{document}
\maketitle

\begin{abstract}
Screening feature selection methods are often used as a preprocessing step for reducing the number of variables before training step.
Traditional screening methods only focus on dealing with complete high dimensional datasets.
Modern datasets not only have higher dimension and larger sample size, but also have properties such as streaming input, sparsity and concept drift.
Therefore a considerable number of online feature selection methods were introduced to handle these kind of problems in recent years. 
Online screening methods are one of the categories of online feature selection methods. 
The methods that we proposed in this research are capable of handling all three situations mentioned above. 
Our research study focuses on classification datasets. Our experiments show proposed methods can generate the same feature importance as their offline version with faster speed and less storage consumption. 
Furthermore, the results show that online screening methods with integrated model adaptation have a higher true feature detection rate than without model adaptation on data streams with the concept drift property. 
Among the two large real datasets that potentially have the concept drift property, online screening methods with model adaptation show advantages in either saving computing time and space, reducing model complexity, or improving prediction accuracy. 
\end{abstract}

\section{Introduction}
With the explosive growth in amount of data available in recent years, the efficiency of data processing stays a heated topic in both hardware and software fields. 
Various approaches were emerged to tackle this challenge, such as cloud computing, batch learning, online learning, more powerful hardware and so on. 
These approaches mainly focus on how to increasing the loading capacity of the systems. 
On the other hand, the screening (filter) feature selection methods, which shows impressive performance in reducing feature dimensions of large high dimensional data and improving overall model performance, especially with ad-hoc implementations, focus on how to scale down the data before considering the loading capacity of the learning system. 
Screening feature selection methods are independent of the learning algorithms, which gives them a fast execution speed. 
They can be ideal add-ons when it comes to improving efficiency without losing performance. 
Built on the foundation of our previous survey on the performance of screening feature selection methods on real large datasets, we dedicate this article to introduce several novel online feature screening methods as extensions of existing offline screening methods to tackle datasets that come from a more modern environment.


In this article, five screening methods are selected from our previous survey \cite{wang2019screening} to be extended to their online versions, sparse versions, and addressing model adaptation issues. 
Among them, T-score\cite{ttestdavis1986statistics} and Fisher Score \cite{fisherduda2012pattern} are mean-variance based methods, while Gini index \cite{gini1912variability}, Chi-square score \cite{liu1995chi2}, and Mutual Information \cite{mutuallewis1992feature} are quantile-based methods. 
Comparative evaluations are then conducted on these methods between their online and offline versions using synthetic and real datasets. 
The model adaptation will also be tested against synthetic data with time-varying features. Finally we will evaluate the performance of these extended methods by showing experimental results on real datasets. 
Given the fact that we don't know which are the true features in real datasets, we will show the performance of these methods by comparing the learners' predictive ability before and after applying screening methods. 
Our complete contribution.....

\textbf{Ours.} Doesn't require previous knowledge of the max/min value of each feature. can stream continues data. Use class label, expendable to multiclass.

\subsection{Related Work}

The study of feature selection on streaming data has a long history. 
Even before the big data boom, there already existed applications that generate and require to process streaming data under  computation and memory constraints. 
One such field is text mining and language processing in topic recognition and spam email identification. 
In cooperating with the well know SpamHunting system, J.R. Méndez et al. introduced a new term selection scheme in \cite{mendez2007relaxing} based on Amount of Information (AI). 
AI is calculated from the appearance of certain terms in the text corpora. It can also be extended as a feature selection method for other types of binary categorical features. This is a method with online criteria calculation and concept adaptation capability. 
Concept adaptation is an essential issue online algorithms need to deal with when processing data streams over a large, possibly infinite, amount of time. 
The fundamental idea is that the relation between the prediction target and its features usually changes over time either gradually or drastically. 
The subset selected by the feature selection methods needs to be adjust to this change as well. Concept adaptation is also called concept drift, feature drift, model adaptation depending on the focus of the dynamic connection between target and features. 

Feature selection methods can be divided into three categories based on their interaction with the learning algorithm: filters, wrappers and hybrids. Filters, also known as screening methods, compute a criterion for each feature and select features according to some selection scheme. They don't involve learner. The aforementioned AI method is a filter method. Wrappers wrap feature selection with any learner. They use a feature importance measure calculated by the learner to decide what features to keep. 
Hybrids, also known as embedded methods, rely on a sparsity inducing built-in regularizer of the learner to drop irrelevant features. One such example is the Lasso\cite{tibshirani1996regression}.

Screening methods have the advantage of fast processing and easy integration with other system. The framework introduced in this article also fall into this category. Considering the large number of streaming high dimensional applications, we feel its advantages make it an appropriate direction to pursue. 

Outside the scope of screening methods, there are other online feature selection methods in the literature. 
In \cite{carvalho2006single} an online learner Modified Balanced Winnow(MBW) was introduced. The authors use the absolute value between the positive weights and negative weights calculated by MBW to rank the features by their importance. 
This is a typical setting for the wrapper methods. Importing the idea of the trade-off between exploration and exploitation, \cite{wang2013online} proposed a method that periodically updates feature weights, and selects feature using an arbitrary classifier. 

Throughout the years, online screening methods have achieved steady development. 
Some of them are introduced to solely handle large data streaming without considering the drift problem. 
Katakis et al. \cite{katakis2005utility} proposed a two stage mechanism for feature selection and data classification. 
They employed a incremental criterion calculation method. 
Features are selected according their rank based on the criterion such as chi-squared score and mutual information. 
Their incremental method is formulated for discrete features, therefore their method can not handle continuous feature. 
In response to the issue of concept drift for streaming data, different strategies were introduced. 

One of the popular methods is sliding window. Its idea is to only consider the most recent data, making it a natural adaptation to the drift problem. 
In \cite{masud2010classification} a chunk structure similar to the sliding window was used by author. 
Deviation weight was introduced in this article as a feature importance measurement and indicator. The feature selection is them perform based on each feature's deviation weight. However the deviation weight is only applicable to categorical features.
Similarly in \cite{gomes2013mining}, an contingency table is maintained for each feature from the sliding window. 
The tables are then used to calculate the criterion such as Mutual information and Chi-squared score and so on. 
The criterion mentioned in their study are only suitable for processing discrete features. 
They also introduced a dynamic threshold to avoid naive selection roles such as top-k features and fixed threshold. 

Another popular strategy widely adapted to handle the drift problem is the fading factor. 
In \cite{sovdat2014updating}, both sliding window and fading factor are used. 
In the fading factor variation, feature measures such as the Gini index and entropy are calculated incrementally over the entire data stream. 
However the criterion they use also works with discrete data only. 
An interesting attempt was presented in\cite{henke2015analysis}, which uses a month as the specific window size for the sliding window to calculate mutual information from categorical features. 
This is very practical the the cases where the drift information is known. 

In some studies, an estimator of criteria is used instead of exact computed criteria mentioned above. 
Keller et al. \cite{keller2015estimating} used a k-nearest neighborhood mutual information estimator introduced by Kraskov el. \cite{kraskov2004estimating} under the sliding window frame. 
The features are then selected based on their mutual information estimators. Instead of calculating some criterion for each feature all the way, in \cite{hammoodi2018real} a concept drift detector detects which features drift first using Feature Velocity and Inter Quartile Range. 
Once a drift is detected, all features that relate to the drift will have their mutual information updated using data in a recent window. 
Then a ranking and selection procedure is carried among all features based on their mutual information.

Despite various situations in which the aforementioned methods can be applied, to our best knowledge, none of the existing online screening methods applies to continues features. 
In order to calculate an appearance based criterion such as mutual information and chi-squared score from continuous features, one needs to be able to discretize the continuous features in an online fashion. 
Many efforts have been put into this area. It has been proven in \cite{munro1980selection} that $O(N)$ space is required for an algorithm to compute the exact quantiles in a single pass of streaming data. 
In order to describe a one pass data stream in a reduced space, approximate quantile calculation methods are needed. 
Some earlier works are \cite{jain1985p} and \cite{Agrawal1995AOS}. Their algorithms are used to calculate uniform quantiles in a single pass. 
In \cite{gama2006discretization} a two-stage framework was proposed to obtain discrete binned data from continuous data. 
The suggested method first constructs many equal width intervals to capture and update the partition of incoming sample points. 
At query time, it aggregates the collected intervals  to generate equal-width histograms or equal-frequency histograms. 
As intuitive as it is, this is a non-deterministic method, in the sense that there are no deterministic guarantees on the estimation error.

Manku et al. introduced a single pass algorithm in \cite{manku1998approximate} to compute a deterministic $\epsilon$-approximate uniform quantile summary. 
It requires prior knowledge of the sample size $N$ and has a space complexity of $O(\frac{1}{\epsilon}\log^2\epsilon N)$. 
Another algorithm that does not require prior knowledge of $N$ was also proposed by Manku et al. in \cite{manku1999random}.  
The space complexity for this algorithm is $O(\frac{1}{\epsilon}(\log^2\frac{1}{\epsilon}+\log^2\log\frac{1}{\delta}))$ with a failure probability of $\delta$.
A more recent approach (the GK algorithm) to compute a deterministic $\epsilon$-approximate quantile summary on a single pass of streaming data without the prior knowledge of $N$ was introduced by Greenwald et al. in \cite{greenwald2001space}. 
This method imposed a tree structure. It is an improvement of Manku's algorithm with a space bound of $O(\frac{1}{\epsilon}\log\epsilon N)$. 
In \cite{zhang2007fast}, an improvement was made on the GK algorithm to significantly reduce the computational cost. 
The computational cost of this multi-level quantile summary algorithm is $O(N\log\frac{1}{\epsilon}\log\epsilon N)$. 
It is shown in their experiments that it can achieve about 200 - 300 $\times$ speedup over the GK algorithm. Its storage requirement of $O(\frac{1}{\epsilon}\log^2\epsilon N)$ is higher than the GK algorithm. 

Recent work in Xgboost\cite{chen2016xgboost} extended \cite{greenwald2001space} and \cite{zhang2007fast} with a focus on processing weighted data. 
This saves space and improves the summary accuracy when dealing data streams containing duplicate values. 
The algorithm introduced in their article has the same guarantee as the GK algorithm. It can be plugged into all GK frameworks and its extensions.

There also exists works that focus on other aspects of calculating quantile summaries. 
In \cite{lin2004continuously} was introduced an algorithms to compute uniform quantiles over sliding windows. 
Cormode et al. \cite{cormode2006space} proposed an algorithm to handle the biased quantile problem. 
It has a storage bound of $O( \frac{\log U}{\epsilon} \log\epsilon N)$ and time complexity of $O(\log \log U)$ where $N$ is the sample stream size and $U$ is the size of the domain from which the sample points are drawn. 
Efforts were also made to compute approximate quantiles for distributed streams and sensor networks. 
Greenwald et al. \cite{greenwald2004power} proposed an algorithm for calculating $\epsilon$-approximate quantiles for sensor network applications. 
In \cite{shrivastava2004medians} an algorithm with space complexity of $O(\frac{1}{\epsilon} \log U)$ was proposed to compute medians and other quantiles in sensor networks.

In this study, we implemented mean-variance based feature screening methods using moving averages. 
We will also introduce quantile based methods based on the weighted quantile summary. 
We extended work in \cite{chen2016xgboost} to generate accurate live bin counts on demand. 
Our proposed algorithm also integrates ways to handle sparse streaming data as well as streaming data featuring concept drift, as an adaptation to the needs of modern applications. 

\section{Proposed Methods}
Two categories of online feature screening methods are introduced in this section. For mean-variance based methods, we will show the concept behind it and illustrate the modifications that are integrated to tackling data stream with sparsity and concept drift. For bin count based methods, We will briefly go over the concept of quantile summary which is the foundation of our introduced methods. Our modification that helps provide exact observation count will also be introduced during the overview. Lastly, we will show the extension we did so that bin count based methods can also tackle data stream having sparsity and concept drift.
 
\subsection{Mean-Variance Based Methods}\label{sec:moveAvgFS}

\subsubsection{Criteria of Interest}

\paragraph{T-score} 
For a feature $x_j$, its T-score is calculated as:
\vspace{5mm}
\begin{equation}
T_j=\frac{|\mu_1-\mu_2|}{\sqrt{\frac{\sigma_1^2}{n_1}+\frac{\sigma_2^2}{n_2}}}\label{eq:T}
\end{equation}
where $\mu_c$, $n_c$, and $\sigma_c$ denote the mean, sample count, and standard deviation of the values of observations belonging to class $c$. The higher the score, the more relevant the feature is to the target variable.

\paragraph{Fisher Score} 
Similarly, the Fisher score of feature $x_j$ is defined as:
\vspace{5mm}
\begin{equation}
Fisher_j=\frac{\sum_{c=1}^Cn_c(\mu_c-\mu)^2}{\sum_{c=1}^Cn_c\sigma_c^2}\label{eq:fisher}
\end{equation}
where $\mu$ is the mean of the feature, and $\mu_c$, $n_c$, and $\sigma_c$ have been defined above. 

\vspace{5mm}
It is clear that equation \eqref{eq:T} and \eqref{eq:fisher} are calculated from basic components such as means and variances. To generalize them to a streaming data setting is quite straightforward.

Without losing generality, we assume that a sample arrives at each time step $t = 1, 2, 3, ..., n$. At time $n$, the running average $\mu_{nj}$ and running mean of squared $MS_{nj}$ of the $j$-th feature that form the sufficient statistics are:
\vspace{5mm}
\begin{align*}
\mu_{1j} &= x_{1j} \\
\mu_{2j} &= \frac{\mu_{1j}}{2}+\frac{x_{2j}}{2} \\
&\vdotswithin{=}   \\
\mu_{nj} &= \frac{n-1}{n}\mu_{(n-1)j}+\frac{1}{n}x_{nj}
\end{align*}
And:
\vspace{5mm}
\begin{align*}
MS_{1j} &= x_{1j}^2 \\
MS_{2j} &= \frac{MS_{1j}}{2}+\frac{x_{1j}^2}{2}  \\
&\vdotswithin{=}   \\
MS_{nj} &= \frac{n-1}{n}MS_{(n-1)j}+\frac{1}{n}x_{nj}^2 
\end{align*}
Consequently the variance of $j$-th feature at time n can be written as:
\vspace{5mm}
\begin{equation}
\sigma_{nj}^2=MS_{nj}-\mu_{nj}^2\\
\end{equation}
Therefore T-score and Fisher Score can incrementally maintain exact calculation as new samples arrive.  

\subsubsection{Sparse Input}
In the scenario of sparse input, since the zero values are no-shown in data stream, one only need to accumulate the running average and running mean of squared  with showed value and keep the record of sample count.

\subsubsection{Model Adaptation}
In order to adapt concept drift, a fading factor strategy is used as a penalty on the history incremented statistics:
\vspace{5mm}
\begin{equation*}
\mu_{nj} = \alpha*\mu_{(n-1)j}+x_{nj} 
\end{equation*}
\begin{equation*}
MS_{nj} = \alpha*MS_{(n-1)j}+x_{nj}^2
\end{equation*}
Where $\alpha$ is a fading factor that takes a user set value in $(0,1)$. 

\paragraph{Sparse Input}
When adaptation is required with sparse input, a time anchor is employed for each feature in every class to recording the last appearance of non-zero values $n_{last}$. It is equal to current sample count. Accumulated statistics then updated the next time a non-zero value appears:
\vspace{5mm}
\begin{equation*}
\mu_{update, j} = \alpha^{(n-n_{last}-1)}*\mu_{last, j}
\end{equation*}
\begin{equation*}
\mu_{nj} = \alpha*\mu_{update, j}+x_{nj} 
\end{equation*}
\begin{equation*}
MS_{update, j} = \alpha^{(n-n_{last}-1)}*MS_{last, j}
\end{equation*}
\begin{equation*}
MS_{nj} = \alpha*MS_{update, j}+x_{nj}^2
\end{equation*}
Where $\mu_{last, j}$ and $MS_{last, j}$ is the penalized running average and running mean of squared at last appearance of a non-zero value.

\subsection{Bin Count Based Methods}\label{sec:quantFS}

\subsubsection{Criteria of Interest}\label{sec:BinCriteria}

\paragraph{Mutual Information} 
For a feature $x_j$, its mutual information can be calculated as:
\vspace{5mm}
\begin{equation}
I(X_j,\by)=\sum_{b=1}^B \sum_{c=1}^C P({x_j}\in{bin}_b,\by=c)\log\frac{P({x_j}\in{bin}_b,\by=c)}{P({x_j}\in{bin}_b)P(\by=c)}
\label{eq:mutual}
\end{equation}
Where $P({x_j}\in{bin}_b,\by=c)$ is the joint probability of having feature values fall into ${bin}_b$ and label value equal to $c$. $P({x_j}\in{bin}_b)$ and $P(\by=c)$ are the marginal probabilities. 

In the case of samples with discrete feature values, the probability can be expressed as:
\vspace{5mm}
\begin{equation*}
P({x_j}\in{bin}_b,\by=c) = \frac{n_{{x_j}\in{bin}_b,\by=c}}{n}
\end{equation*}
\begin{equation*}
P({x_j}\in{bin}_b) = \frac{n_{{x_j}\in{bin}_b}}{n}
\end{equation*}
\begin{equation*}
P(\by=c) = \frac{n_{\by=c}}{n}
\end{equation*}
Where $n$, $n_{{x_j}\in{bin}_b,\by=c}$, $n_{{x_j}\in{bin}_b}$, $n_{\by=c}$ denote the sample count that fall into respective value groups.

\paragraph{Chi-squared Score} 
With a similar definition of $n$'s, the chi-squared score of a feature $x_j$ can be defined as:
\vspace{5mm}
\begin{equation}
\chi^2_j=\sum_{b=1}^B\sum_{c=1}^C\frac{(n_{{x_j}\in{bin}_b,\by=c}-\hat{n}_{{x_j}\in{bin}_b,\by=c})^2}{\hat{n}_{{x_j}\in{bin}_b,\by=c}}\label{eq:chi2}
\end{equation}
Where:
\vspace{5mm}
\begin{equation*}
 \hat{n}_{{x_j}\in{bin}_b,\by=c}=\frac{n_{{x_j}\in{bin}_b}n_{\by=c}}{n}
 \end{equation*}

\paragraph{Gini Index} 
For a given feature $x_j$, let ${A}_{h}=\{i, x_{ij}\leqslant h\}$ denote the number of samples whose values of the $j$-th feature is smaller than or equal to $h$ and ${B}_{h}=\{i, x_{ij}> h\}$. Its Gini Index can be expressed as:
\vspace{5mm}
\begin{equation}
Gini_j=P(A_h)(1-\sum_{c=1}^CP(C_c|A_h)^2)+P(B_h)(1-\sum_{c=1}^CP(C_c|B_h)^2)\label{eq:gini}
\end{equation}
Where $P({A}_{h})$ is the number of samples in subset ${A}_{h}$ divided by the number of total samples. $P({C}_{c}|{A}_{h})$ is the conditional probability of samples having label $c$ given that they are in subset ${A}_{h}$. Let $n_{{x_j}\in{A}_h,\by=c}$ denote the number of samples in ${A}_{h}$ with label $c$. Let $n_{{x_j}\in{A}_h}$ denote the number of samples in ${A}_{h}$. Then $P({C}_{c}|{A}_{h})$ can be calculated as $n_{{x_j}\in{A}_h,\by=c}$/$n_{{x_j}\in{A}_h}$. The same goes for $P({B}_{h})$ and $P({C}_{c}|{B}_{h})$.
$h$ is chosen to give the minimum Gini Index for each feature.

\vspace{5mm}
In order to compute aforementioned criterion from data stream with continuous values. A proper online discretization method must be applied. Therefore based on the $\epsilon$-approximate quantile summary structure in \cite{greenwald2001space}, \cite{zhang2007fast} \cite{chen2016xgboost}, we introduce an improved implementation algorithm to generate an on demand bin count from data stream with continuous values. Furthermore, we extend our algorithm to adapt to sparse input and concept drift scenario. In the following segment we will first present an overview of  $\epsilon$-approximate quantile summary structure and its operation. Then we will illustrate our extended methods in detail.

\subsubsection{Quantile Summary} \label{sec:quantileS}
The basic idea is to use several container like sub-summaries $s$ to store approximated ranking information of partial data stream. In turn, an aggregated summary $S$ of sub-summaries can describe the entire data stream. Various operation are conducted periodically to maintain the estimations in these containers such that at any time $n$, the summary $S(n)$ can answer any $r$-quantile query with $\epsilon n$ precision.

A sub-summary consists of several tuples $s=\{T_1,T_2,...,T_b\}$. 
Each tuple in the form of T=$(v,\tilde{r}^-,\tilde{r}^+,\tilde{w})$ describes a  number of similar data points from the data stream. 
$v$ denotes the data value estimation this tuple represents. $\tilde{r}^-$ and $\tilde{r}^+$ are the lowest and highest estimated ranks of this value in the current sub-summary. 
$\tilde{w}$ is the accumulated weight value which represents how many data points this tuple covers. 
The sub-summary $s$ is sorted by the values $v$. 
Given a small input stream $Q=\{(x_1,w_1),(x_2,w_2),...,(x_n,w_n)\}$, where $(x_i,w_i)$ is a data point. $x_i$ denotes its value and $w_i$ is its weight. Usually data weights are set to $1$. For each tuples, two rank functions and a weight function are defined.
\begin{equation}
r^-(v)=\sum_{(x,w)\in Q,x<v}w
\label{eq:rmin}
\end{equation}
\begin{equation}
r^+(v)=\sum_{(x,w)\in Q,x\leqslant v}w
\label{eq:rmax}
\end{equation}
\begin{equation}
w(v)=r^+(v)-r^-(v)=\sum_{(x,w)\in Q,x=v}w
\label{eq:weight}
\end{equation}

The weight of the entire sub-summary is defined as:
\begin{equation}
w(s)=w(Q)=\sum_{(x,w)\in Q}w
\label{eq:totalweight}
\end{equation}

Without loss of generality, in the rest of this article, weight will be used to refer to the calculated weight value of $w(v)$. Rank will be used to refer to the calculated rank value of $r^+(v)$ or $r^-(v)$.

\begin{figure}[htb]
\centering
\includegraphics[width=0.8\linewidth]{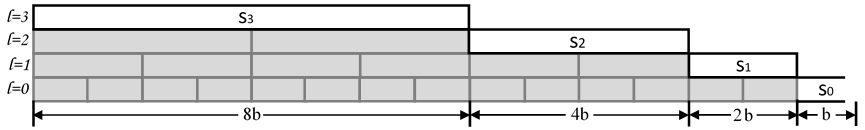}
\caption{Multi-level summary \cite{zhang2007fast}: The length of $s_l$ in the figure represents its coverage of data points. $s_0$ contains the summary for the most recent data input block. $s_l$ consists the summary of the oldest $2^l$ data blocks. At each level, $s_l$ is maintained as an $\epsilon_l$-summary.}\label{fig:multiLevel}
\end{figure}

Summary $S$ is consist of a multi-level sub-summary structure, shown in Figure \ref{fig:multiLevel}, when it is not queried. It is used to maintain the desired precision as well as speed up the calculation. L is the total number of levels. $s_l$ denotes the sub-summary at level $l, l=0, 1,..., L$. The whole data stream is divided into consecutive segments of size $b=\ceil[\bigg]{\frac{L}{\epsilon}}$, where $L$ is the largest integer that makes $b2^{(L-1)}\leqslant N$. 

At the lowest level, $s_0$ is defined to hold all recently arrived points until it researches size $b$. The tuples in $s_0$ are constructed with
\begin{equation}
v=x_i, \tilde{r}^-(x_i)=r^-(x_i), \tilde{r}^-(x_i)=r^-(x_i), \tilde{w}(x_i)=w(x_i)
\label{eq:estTuple}
\end{equation}
Therefore $s_0$ is a 0-approximate summary. It can answer all query questions exactly. Algorithm \ref{alg:quantileS} shows the basic procedures of summarizing when a new data point $x_i$ in the data stream arrives. In the algorithm, $s_{temp}$ denotes a temporary sub-summary. $s_l$ denotes the $l$th level in the multi-level summary structure mentioned above.

\begin{algorithm}[htb]
   \caption{{\bf General Procedure for Quantile Summary}}
   \label{alg:quantileS}
\begin{algorithmic}
   \STATE {\bfseries Input:} $x_i$, where i=1, ..., t
\end{algorithmic}
\begin{algorithmic} [1]
\STATE push $x_i$ into $s_0$
\IF {$size(s_0) < b$}
	\STATE  go back to line 1
\ELSE
\STATE $s_{temp}$=PRUNE($s_0,\frac{size(s_0)}{2}$)
\STATE clear $s_0$
\FOR {$l = 1, ..., L$}
\IF {$size(s_l)$ is 0}
\STATE $s_l=s_{temp}$
\STATE clear $s_{temp}$; break
\ELSE
\STATE $s_{temp}$=MERGE($s_{temp},s_l$)
\STATE $s_{temp}$=PRUNE($s_{temp}$,$\frac{size(s_{temp})}{2}$)
\IF {$size(s_{temp}) < b$}
\STATE $s_l=s_{temp}$; clear $s_{temp}$; break
\ELSE
\STATE clear $s_l$
\ENDIF
\ENDIF
\ENDFOR
\ENDIF

\STATE {\bfseries Output:}  S=MERGE($s_0, s_1, ..., s_L$)

\end{algorithmic}
\end{algorithm}

PRUNE($s,\frac{k}{2}$) is an operation that converts a sub-summary $s$ with size $k$ and precision $\epsilon_0$ into a sub-summary with size $\frac{k}{2}+1$ and precision $\epsilon_0+\frac{1}{k}$. It is shown in \cite{zhang2007fast} that each level in the summary can maintain an error less than $\epsilon$. 

In the PRUNE operation, maximum $g$ tuples are chosen from the input sub-summary according to position indicator $d=\frac{i-1}{g}w(s), i=1, 2,..., b+1$. Algorithm \ref{alg:pickx} shows how to use the query operation $Q(s,d)$ to choose tuples to form the new sub-summary. $\tilde{r}^-(x_i), \tilde{r}^-(x_i), \tilde{w}(x_i)$ of the selected tuples are copied from the original sub-summary.

\begin{algorithm}[htb]
   \caption{{\bf Query Function Q(s,d)}}
   \label{alg:pickx}
\begin{algorithmic}
   \STATE {\bfseries Input:} d: $0\leqslant d\leqslant w(s)$, s is a sub-summary with tuple value $v_i$ = $x_i$, $i = 1, 2, ..., k$
\end{algorithmic}
\begin{algorithmic} [1]
\IF {$d<\frac{1}{2}[\tilde{r_s}^-(x_1)+\tilde{r_s}^+(x_1)]$}
	\STATE  return $x_1$
\ENDIF
\IF {$d\geqslant\frac{1}{2}[\tilde{r_s}^-(x_k)+\tilde{r_s}^+(x_k)]$}
	\STATE  return $x_k$
\ENDIF
\STATE  Find i such that
\STATE $\frac{1}{2}[\tilde{r_s}^-(x_i)+\tilde{r_s}^+(x_i)]\leqslant d <\frac{1}{2}[\tilde{r_s}^-(x_{i+1})+\tilde{r_s}^+(x_{i+1})]$
\IF {$2d<\tilde{r_s}^-(x_i)+\tilde{w_s}(x_i)+\tilde{r_s}^+(x_{i+1})-\tilde{w_s}(x_{i+1})]$}
	\STATE  return $x_i$
\ELSE
	\STATE  return $x_{i+1}$
\ENDIF
\end{algorithmic}
\end{algorithm}

Another operation is MERGE. MERGE($s_1$,$s_2$) combines two sub-summaries into one sub-summary.  Tuples from two sub-summaries $s_1$ and $s_2$ are sorted together by their value $v$. 
For each unique value $v$ from $s_1$ and $s_2$, $\tilde{r}^-(x_i), \tilde{r}^-(x_i), \tilde{w}(x_i)$ are updated as follows
\begin{equation}
\tilde{r}^-(v)=\tilde{r_{s_1}}^-(v)+\tilde{r_{s_2}}^-(v)
\label{eq:uprmin}
\end{equation}
\begin{equation}
\tilde{r}^+(v)=\tilde{r_{s_1}}^+(v)+\tilde{r_{s_2}}^+(v)
\label{eq:uprmax}
\end{equation}
\begin{equation}
\tilde{w}(v)=\tilde{w_{s_1}}(v)+\tilde{w_{s_2}}(v)
\label{eq:upweight}
\end{equation}

Let the precisions of two sub-summaries before merging to be $\epsilon_a$ and $\epsilon_b$. The precision of merged sub-summary $s_{temp}$ is $\max(\epsilon_a,\epsilon_b)$ \cite{chen2016xgboost}. MERGE operation can also be apply to more than two sub-summaries. 
It is shown, at the end of Algorithm \ref{alg:quantileS}, the aggregated summary $S$ is the out come of MERGE operation over all sub-summaries. Therefore, $S$ in its result form has the same structure as $s$. 

It is shown in \cite{zhang2007fast} that the outcome summary $S$ is an $\epsilon$-approximate summary of the entire stream. 

According to Algorithm \ref{alg:quantileS}, whether to perform a PRUNE or MERGE operation is based on the segment size $b$ which is determined by the stream size $N$. In order to obtain a summary from a data stream with stream size $N$ unknown, the input data stream is divided into pieces of disjoint sub-streams $B_i, i=0, 1,..., m$.
$B_i$ has size of $\frac{2^i}{\epsilon}$ and covers data arriving in the time interval $[\frac{2^i-1}{\epsilon},\frac{2^{i+1}-1}{\epsilon})$. With fixed sub-stream size aggregated summaries can be obtained. 
Due to the fact that the output of $S$ and $s$ have the same structure, $S$ can too perform PRUNE and MERGE operation with other $S$. Therefore after obtaining $S_i$ for each sub-stream $B_i$, a multi-level summary structure $[S]$ now can be constructed from the summary of each sub-stream $S_i$. The procedure is illustrated below.
\begin{enumerate}
\item The summary $S_c$ of current sub-stream $B_c$ is updated and maintained until the last data point in $B_c$ has arrived. $\epsilon'=\frac{\epsilon}{2}$ is used to set size limit. 
\item A $\frac{\epsilon}{2}$-approximate summary is obtained as an output of $S_c$. The output is then set to PRUNE with the desired size of $\frac{2}{\epsilon}$ and assigned to $S_i$.
\item A set of summaries of all sub-streams $\widetilde{S}=\{S_0, S_1,..., S_m\}$ is computed. $[S]$ is obtained by MERGE summaries in $\widetilde{S}$.
\end{enumerate}

\subsubsection{Exact Weight Update} \label{sec:bincount}
When used as a query algorithm, only the rank $r$ and value $v$ of the tuple will be used to answer the question. According to the original PRUNE operation in section \ref{sec:quantileS}, the selected tuples along with their stored elements are directly moved to the resulting summary. Although this behavior causes the lost of half of the elements that store weight values. It still guarantees the $\epsilon$ maximum error as shown in \cite{chen2016xgboost} \cite{greenwald2001space} \cite{zhang2007fast}. However, to the end of providing accurate bin count as inputs to these bin count based criteria, each tuple is treated as a mini-bin. The weight $w$ of each tuple will be used to calculate the final bin count. Therefore, the preservation of the complete weight values is required. 

In the PRUNE operation, after each selection, instead of carrying elements directly  from original tuples to tuples in new summary, different procedures are taken. 
Let the value of each tuple associate with original summary be $v_k$, $k = 1, 2, 3, ..., i, ..., j, ..., b$. $i$ is the index for last selected tuple. $j$ is the index of currently selected tuple. Let the value of each tuple associate with output summary be $u_h$, $h = 1, 2, 3, ..., q, ..., b/2$. $u_q$ is the corresponding tuple derived from $v_j$. Following adjustments are made:
\begin{align*}
u_q&=v_j\\
\tilde{r}^-(u_q)&=\tilde{r}^-(v_{i+1})\\
\tilde{r}^+(u_q)&=\tilde{r}^+(v_j)\\
\tilde{w}(u_q)&=\sum_{k=i+1}^j\tilde{w}(v_k)
\end{align*}
The definition of $\epsilon$-approximate quantile summary states that let $Q$ be the set of all data points covered by the summary, if for any $x\in Q$
\begin{equation}
\tilde{r}^+(x)-\tilde{r}^-(x)-\tilde{w}^+(x)\leqslant\epsilon w(Q)
\end{equation}

The above adjustment leads the left hand side of the inequality to zero which satisfied the requirement of a $\epsilon$-approximate quantile summary.

\subsubsection{Quantile Binning} \label{sec:binning}
With the weight of each data point set to 1, the weight element in a tuple can represent the number of data points this tuple covered. Using procedure in last two sections, a final summary $[S]$ consists of $m$ tuples is generated. $[S]$ is then further aggregated into smaller number of denser bins. For the purpose of comparability with our offline method survey \cite{wang2019screening}, procedures that mimic the discretization in \cite{discretization} are introduced. Details are shown in Algorithm \ref{alg:agg}. $d_{inter}$ denotes the interval length when data points are equally divided into $K$ segments. $Bin_k$ denotes the bin count in $k$th Bin.   $p_i$ is the position index of $i$th cutoff point. $T_j.w$ represents the weight of the $j$th tuple.


\begin{algorithm}[htb]
   \caption{{\bf Bin Aggregation Procedure}}
   \label{alg:agg}
\begin{algorithmic}
   \STATE {\bfseries Input:} $[S]$ = {$T_1, T_2, ..., T_m$}, $N$: number of total data so far, $K$: user defined number of final bins.
\end{algorithmic}
\begin{algorithmic} [1]
\STATE $d_{inter} = \floor*{\frac{N}{K}}$, h = 0
\FOR {$i = 1, ..., K-1$} 
	\STATE  $p_i = i*d_{inter}$
\ENDFOR

\FOR {$k = 1, ..., K$} 
	\STATE  $Bin_k = 0$
\ENDFOR

\FOR {$j = 1, ..., m$} 
	\STATE  $Bin_h = Bin_h + T_j.w$
	\IF {$|Bin_h| \geqslant p_h$}
		\STATE $h_{temp} = h$
		\STATE h = h + 1
		\WHILE {$|Bin_{h_{temp}}| \geqslant p_{h}$}
			\STATE  h = h + 1
		\ENDWHILE
	\ELSE
		\STATE continue
	\ENDIF
\ENDFOR
\STATE {\bfseries Output:}  $Bin_k, k = 1, 2, ..., K$
\end{algorithmic}
\end{algorithm}
Criteria score in section\ref{sec:BinCriteria} can then be calculated using $Bin_k$

\subsubsection{Sparse Input} 
Different from using running sum for mean-variance based methods, the zero values are assigned weight one. Therefore in the scenario of sparse input, zero values need to be processed through the summary. Processing zero values every time one shows up will be computationally inefficient. We take the advantage of the fact that sparse data has enormous zero values and quantile summary directly stacks the weights of identical data values together. The total number of zero values are recorded. When the algorithm is called to provide feature importance score, a single data point $(x,w)$, with $x$ equals to 0 and $w$ equals to the recorded number, is pushed through the algorithm before aggregating the summary.

\subsubsection{Model Adaptation}
In the situation of concept drift, a fading factor strategy is still adopted. Let $\alpha$ denote the fading factor, $w_i$ be the weight value of data point at $i$ time and $W_i$ be the weight values in the multi-level quantile summary at $i$ time. 

\begin{equation}\label{eq:summarydrift}
W_i = \alpha W_{i-1} + w_i
\end{equation}

Notice in equation \ref{eq:summarydrift}, the weight in $W_{i-1}$ needs to be updated every time a new data point arrive. This includes all the tuples on all the levels in the summary. Such high frequency repeatedly calculation is very time consuming. Therefore we split update process into two parts. The first part, we conduct update only on most recently established tuples in $s_0$(see Fig \ref{fig:multiLevel} to refresh memory of $s_0$). The second part, based on the number of data points covered in $s_0$, we only penalize the weights in multi-level structure whenever a PRUNE and MERGE operation initialized by $s_0$ happens. Let the number of data points in $s_0$ be $k$ and the weights in multi-level summary be $W_i$. The update for part two become:
\begin{equation}
W_i = \alpha^k W_{i-1}
\end{equation}

\paragraph{Sparse Input} When sparse inputs are received in a model adaptation setting, since the weight of each data point is penalized according to the order that it arrives, injecting the weights of all zero value data points at the end can not provide the correct penalized weight. In order to keep weight accumulation matching with data point order for the zero value data points, time anchors and universal weight maps are employed. 
The universal weight map is a vector that store the accumulated penalized weight at each timestamp for each class. These maps are used across all features. 
The time anchors are designed to record the timestamp of the last non-zero data point. 

When incoming data stream has greater or equal to two class. Simply recording the class label and recover the penalized weight for zero values by repeatedly penalizing the weights in summary so far is extremely time consuming. On the contrary,  universal weight map and time anchor can achieve the same purpose with very small extra storage space and faster computing speed without trigger inter-class penalization complex. Given universal weight map recorded up until now for each class. The weight for all zero values between last non-zero data point and current non-zero data point can be calculated. Let time indices for last non-zero data point be $a$ and current non-zero data point be $b$.
\begin{equation}
\bw_c = M_{b, c} - M_{a, c}*\alpha^{(b-a)}
\end{equation}
$\bw_c$ denotes the recovered weight for zero values. $M_{b, c}$ indicates the recorded weight value for $c$ class at time index $b$. When a non-zero value arrives, $\bw_c$ is calculated and added to summary before any other procedures.

\subsection{Minibatch} \label{sec:minibatchFS}
It is noticed during our experiments that for non-spare inputs, the summaries of all features are required to be visited. This behavior builds increased computing time. 
Therefore within a reasonable storage budget, batch data handling are integrated with the algorithm, as it gives a considerably acceleration to the algorithm by reducing the visit frequency. This batch procedure is denoted by minibatch in the rest of the text.

\section{Evaluation of Proposed Methods}
\subsection{Online-Offline Methods Comparison}\label{sec:onoff}

To perform the experiments, each dataset is processed in a one-pass fashion by both online and offline version of screening methods respectively. 
Each dataset is passed through each algorithm and parameter setting once and the weight scores for all features are calculated. 
For quantile summary based methods, we use $K=5$ quantile bins throughout all our experiments. 
We test the computation time by fixing either the minibatch size or the precision parameter $\epsilon$ and varying the other parameter. 
When testing  the approximation accuracy, we only fix the minibatch size. The fixed values are 250 for the minibatch and $0.001$ for $\epsilon$. 
The varying ranges are $\epsilon=\frac{1}{f}$, where $f\in \{ 5, 50, 100, 500, 1000, 1500, 2000\}$ and minibatch$=2^k$, where $k\in \{0, 1, 2, ..., 11\}$. 
The feature rank is calculated by sorting the feature weights monotonically according to importance score obtained by the corresponding screening method. 
Features that have low rank value are more important (i.e. feature with rank value 1 is the most important). 
For Gini index, the feature that has smaller weight has lower (better) rank value. 
On the contrary, for  the other methods, the feature that has larger weight has lower rank value. 
Feature weights and feature ranks were used to construct different kinds of tables to evaluate the performance of online methods compared to offline methods.


\subsubsection{Construction of Comparison Tables}
Five types of tables are constructed according to minibatch size, feature weights and feature rank.

\noindent 1) {\em The influence of minibatch size on computation time.} This table shows the computation time (in milliseconds) of the online quantile compared to the offline quantile when varying the minibatch size. A minibatch size of one is equivalent to an online method without minibatch processing.

\noindent 2) {\em The influence of the precision parameter $\epsilon$ on computation time.}  This table shows the computation time (in milliseconds) of the online quantile compared to the offline quantile when varying the value of $\epsilon$. 

\noindent 3) {\em The influence of $\epsilon$ on count differences per feature per bin.} This table shows the average count differences between online quantile and offline quantile when varying the value of $\epsilon$. Each cell gives the average count difference per feature per bin.

\noindent 4) {\em The influence of $\epsilon$ on score accuracy.} This table shows the mean score difference ratio between online methods and offline methods when varying the value of $\epsilon$. 
The mean score difference ratio is calculated by 
\begin{equation}
\text{DR}=\frac{1}{p}\sum_{j=1}^p{\frac{|w_{on}^j-w_{off}^j|}{\max(W_{off})-\min(W_{off})}}, \label{eq:accuracy}
\end{equation}
 where $p$ is the total number of features. 
$W_{on}$ and $W_{off}$ are the score vectors of all features generated from online and offline methods. $w_{off}^j$ and $w_{off}^j$ are the scores of the $j$-th feature from online and offline methods.

\noindent 5) {\em The influence of $\epsilon$ on the rank accuracy among top 10\% most important features.} This table shows the average unmatched ranking ratio of online methods with respect to corresponding offline methods when varying the value of $\epsilon$. 
Only the top 10\% of features ranked by offline scores are involved. 
The average mis-rank ratio is calculated as $\frac{\sum_{j=1}^p{r_{on}^j\neq r_{off}^j}}{p}$, where $r_{off}^j$ and $r_{off}^j$ are the rank values of the $j$-th feature from online and offline methods.
%
%





\subsubsection{Data Sets}
Datasets used for online-offline methods comparison are shown in Table \ref{tab:dataset2}.

\begin{table}[t]
\vskip -1mm
\begin{center}
\caption{The datasets used for evaluating the online/offline screening methods.}\label{tab:dataset2}
\begin{tabular}{|l|c|c|c|c|}
\hline
Dataset &\thead{Learning type} &\thead{Feature type} &\thead{Number of\\ features} &\thead{Number of\\ observations}\\[0.5ex]
\hline
\hline
\href{http://featureselection.asu.edu/datasets.php}{Gisette} \cite{guyon2005result} &Classification &Continuous &5000 &7000\\[0.5ex]
\hline
\href{https://archive.ics.uci.edu/ml/datasets/dexter}{Dexter} \cite{guyon2005result} &Classification &Continuous &20000 &600\\[0.5ex]
\hline
\href{http://featureselection.asu.edu/datasets.php}{Madelon} \cite{guyon2005result} &Classification &Continuous &500 &2600\\[0.5ex]
\hline
\href{http://featureselection.asu.edu/datasets.php}{SMK\_CAN\_187} \cite{spira2007airway} &Classification &Continuous &19993 &187\\[0.5ex]
\hline
\href{http://featureselection.asu.edu/datasets.php}{GLI\_85} \cite{freije2004gene} &Classification &Continuous &22283 &85\\[0.5ex]
\hline
\href{https://archive.ics.uci.edu/ml/datasets/dorothea}{Dorothea} \cite{guyon2005result} &Classification &Continuous &100,000 &1,150\\[0.5ex]
\hline
Url \cite{ma2009identifying}&Classification &Continuous, Binary &74,110 &16,000\\[0.5ex]
\hline
\href{https://www.csie.ntu.edu.tw/~cjlin/libsvmtools/datasets/binary.html}{Kdd12} \cite{juan2016field}&Classification &Binary &48,957 &16,000 \\[0.5ex]
\hline
\hline
\end{tabular}
\end{center}
\vskip -6mm
\end{table}

Gisette, Dexter, Madelon are part of the NIPS 2003 Feature selection challenge \cite{guyon2005result} and are also available on the UCI Machine Learning Repository. 
Dorothea is part of the NIPS 2003 Feature selection challenge and is also available on the UCI Machine Learning Repository. 
We combined the training set and validation set in order to get a larger sample body. 
The Url dataset contains a total of 121 data files, one for each monitored day. We only used data from Day0 in our experiments. 
The Kdd12 dataset originates from the second track of the KDD Cup 2012. 
The raw version can be found on kaggle.com, made available by the organizers and Tencent Inc. 
The data we use comes from LIBSVM \cite{CC01a}: a library for support vector machines. Only the first 16000 samples were used in our experiments.

\subsubsection{Results}
The following results are based on the output generated using Matlab 2018b \cite{Matlab2018b}. 
For the offline screening methods we used the same Matlab 2018b implementations as those in \cite{wang2019screening}. 
The online screening methods were implemented by ourselves.

\paragraph{Comparison of the Moving Average Based Methods}

It is shown in Section \ref{sec:moveAvgFS} that the moving average based online screening methods can achieve exactly the same result as their offline version. Therefore we don't provide any real data analysis here.

\paragraph{Comparison of the Online Quantile Based Methods}

First we conducted a study of the computation time of the online quantile based methods. 
Here we only evaluate the time from when the data was input data to when the sample points counts in each bin were returned.
We first conducted an experiment to evaluate the relation between computation time and minibatch size to estimate the minibatch size that we were going to use to obtain an acceptable speed in our following experiments. 
In this experiment, the $\epsilon$ was fixed at $0.001$. The minibatch size was varied as $2^k$, where $k = 0, 1, 2, ..., 11$. 
The result is shown in Table \ref{tab:TimeMini}, where the time is measured in milliseconds. 
Compared across all datasets, it can be seen that the computation times are gradually stable for minibatch sizes lager than 128, with an acceptable fluctuation. 
For most datsets, the computation speed of the online quantile outperforms that of the offline quantile long before the minibatch size reaches 128. 
$SMK\_CAN\_187$ has only 187 observations. There the minibatch doesn't improve computation time after minibatch size 128. 
As a result 250 was selected as minibatch size in all of the following studies.

\begin{table}[t]
\vspace {-1mm}
\begin{center}
\caption{Influence of minibatch size on computation time for online quantile compared to offline quantile.}
\label{tab:TimeMini}
\begin{tabular}{|l|S|S|S|S|S|S|S|}\hline%
minibatch&{url} &{$SMK\_CAN\_187$} &{dexter} &{dorothea} &{gisette} &{kdd12} &{madelon}\\\hline\hline
\csvreader[late after line=\\\hline]%
{TableOfRuntime_mini.csv}{minibatchSize=\minibatch,url=\url,SMK_CAN_187=\SMK,dexter=\dexter,dorothea=\dorothea,gisette=\gisette,kdd12=\kdd,madelon=\madelon}
{\minibatch&\url&\SMK&\dexter&\dorothea&\gisette&\kdd&\madelon}%
\end{tabular}
\end{center}
\vspace{-5mm}
\end{table}

\begin{table}[htb]
\vspace {-2mm}
\begin{center}
\caption{Influence of $\epsilon$ on computation time for online quantile compared to offline quantile.}
\label{tab:TimeEps}
\vskip -2mm
\begin{tabular}{|l|S|S|S|S|S|S|S|}\hline
epsilon&{url} &{$SMK\_CAN\_187$} &{dexter} &{dorothea} &{gisette} &{kdd12} &{madelon}\\\hline\hline
\csvreader[late after line=\\\hline]%
{TableOfRuntime_eps.csv}{epsilon=\epsilons,url=\url,SMK_CAN_187=\SMK,dexter=\dexter,dorothea=\dorothea,gisette=\gisette,kdd12=\kdd,madelon=\madelon}
{\epsilons&\url&\SMK&\dexter&\dorothea&\gisette&\kdd&\madelon}%
\end{tabular}
\end{center}
\vspace{-5mm}
\end{table}
The next experiment was conducted to verify that in a real data scenario, computation time increases as $\epsilon$ decreases. 
In Table \ref{tab:TimeEps} is shown that it is certainly the relation between computation speed and $\epsilon$. 
Furthermore, in some case where the datasets have large sample size e.g. url and kdd12, a low $\epsilon$ value can cause the online methods to use more computation time than offline methods. 
To improve the computation speed, larger minibatch size ought to be used. 
Here in order to maintain the consistency, we still use 250 as minibatch size.
\begin{table}[t]
\vspace {-1mm}
\begin{center}
\caption{Influence of $\epsilon$ on count differences between online and offline quantiles (results are reported per feature per bin).}
\label{tab:CountEps}
\begin{tabular}{|l|c|c|c|c|c|c|c|}\hline%
epsilon&url &$SMK\_CAN\_187$ &dexter &dorothea &gisette &kdd12 &madelon\\\hline\hline
\csvreader[late after line=\\\hline]%
{TableOfCDRatio_eps.csv}{epsilon=\epsilons,url=\url,SMK_CAN_187=\SMK,dexter=\dexter,dorothea=\dorothea,gisette=\gisette,kdd12=\kdd,madelon=\madelon}
{\epsilons&\url&\SMK&\dexter&\dorothea&\gisette&\kdd&\madelon}%
\end{tabular}
\end{center}
\vspace{-5mm}
\end{table}

The following experiments are regarding accuracy. In Table \ref{tab:CountEps} is shown the count differences between online and offline quantile methods when the $\epsilon$ value was varied. 
The results are reported as sample count differences per bin per feature. Although different datasets have different schedule, all differences were annihilated after $\epsilon$ decreases below $0.001$. 
The online quantile method can in practice achieve the same counts per bin as the offline quantile when $\epsilon$ is lower than a certain level. 
Cross checking with Table \ref{tab:TimeEps} and Table \ref{tab:TimeMini}, we can see that online methods can still achieve speed advantage with little to no error in bin count estimation.

\begin{table}[htb]
\vspace {-1mm}
\begin{center}
\caption{Influence of $\epsilon$ on accuracy \eqref{eq:accuracy} of online chi-square/gini index/mutual information scores compared to their offline versions}
\label{tab:WeightEps}
\begin{tabular}{|l|c|c|c|c|c|c|c|}\hline%
epsilon&url &$SMK\_CAN\_187$ &dexter &dorothea &gisette &kdd12 &madelon\\\hline
\multicolumn{8}{l}{Chi-square}\\
\hline
\csvreader[late after line=\\\hline]%
{TableOfMAR_eps_chi2.csv}{epsilon=\epsilons,url=\url,SMK_CAN_187=\SMK,dexter=\dexter,dorothea=\dorothea,gisette=\gisette,kdd12=\kdd,madelon=\madelon}
{\epsilons&\url&\SMK&\dexter&\dorothea&\gisette&\kdd&\madelon}%
\hline
\multicolumn{8}{l}{Gini index}\\
\hline
\csvreader[late after line=\\\hline]%
{TableOfMAR_eps_gini.csv}{epsilon=\epsilons,url=\url,SMK_CAN_187=\SMK,dexter=\dexter,dorothea=\dorothea,gisette=\gisette,kdd12=\kdd,madelon=\madelon}
{\epsilons&\url&\SMK&\dexter&\dorothea&\gisette&\kdd&\madelon}%
\hline
\multicolumn{8}{l}{Mutual information}\\
\hline
\csvreader[late after line=\\\hline]%
{TableOfMAR_eps_mi.csv}{epsilon=\epsilons,url=\url,SMK_CAN_187=\SMK,dexter=\dexter,dorothea=\dorothea,gisette=\gisette,kdd12=\kdd,madelon=\madelon}
{\epsilons&\url&\SMK&\dexter&\dorothea&\gisette&\kdd&\madelon}%
\end{tabular}
\end{center}
\vspace{-5mm}
\end{table}
Table \ref{tab:WeightEps} shows the accuracy measure from Eq. \eqref{eq:accuracy} of the scores calculated by three online methods compared to their offline counterparts when the $\epsilon$ value was varied. 
The scores were computed using the Chi-square score, Gini index and Mutual information respectively. 
Among the three methods Gini index provides slightly better accuracy than the other two methods. 
All of them can achieve zero error when $\epsilon$ is smaller than 0.001. 
The results shown here align with the count accuracy shown in Table \ref{tab:CountEps}.


Finally, we check the ranking accuracy of each of the three online methods with respect to their offline version. 
In Table \ref{tab:RankEps} is shown that the ranking error can be eliminated when using $\epsilon$ smaller than 0.01, even for the most difficult datasets. 
The features compared here are top 10\% ranked features according to scores calculated by the offline methods. 
10 percent of all features is usually the zone where most important features reside. 
Therefore, for the sole purpose of screening feature selection, using online methods with an relative large $\epsilon$ is enough to assure that we can get identical feature ranking results as the offline methods.

\begin{table}[htb]
\vspace {-1mm}
\begin{center}
\caption{Influence of $\epsilon$ on unmatched ranking among top 10\% features for chi-square/gini index/mutual information score according to its offline feature ranking}
\label{tab:RankEps}
\begin{tabular}{|l|c|c|c|c|c|c|c|}\hline%
epsilon&url &$SMK\_CAN\_187$ &dexter &dorothea &gisette &kdd12 &madelon\\\hline
\multicolumn{8}{l}{Chi-square}\\
\hline
\csvreader[late after line=\\\hline]%
{TableOfMisRank10percent_eps_chi2.csv}{epsilon=\epsilons,url=\url,SMK_CAN_187=\SMK,dexter=\dexter,dorothea=\dorothea,gisette=\gisette,kdd12=\kdd,madelon=\madelon}
{\epsilons&\url&\SMK&\dexter&\dorothea&\gisette&\kdd&\madelon}%
\hline
\multicolumn{8}{l}{Gini index}\\
\hline
\csvreader[late after line=\\\hline]%
{TableOfMisRank10percent_eps_gini.csv}{epsilon=\epsilons,url=\url,SMK_CAN_187=\SMK,dexter=\dexter,dorothea=\dorothea,gisette=\gisette,kdd12=\kdd,madelon=\madelon}
{\epsilons&\url&\SMK&\dexter&\dorothea&\gisette&\kdd&\madelon}%
\hline
\multicolumn{8}{l}{Mutual information}\\
\hline
\csvreader[late after line=\\\hline]%
{TableOfMisRank10percent_eps_mi.csv}{epsilon=\epsilons,url=\url,SMK_CAN_187=\SMK,dexter=\dexter,dorothea=\dorothea,gisette=\gisette,kdd12=\kdd,madelon=\madelon}
{\epsilons&\url&\SMK&\dexter&\dorothea&\gisette&\kdd&\madelon}%
\end{tabular}
\end{center}
\vspace{-5mm}
\end{table}
\subsection{Online Screening Methods with Model Adaptation}\label{sec:syn}
In his section, online screening methods with model adaptation capability are evaluated for their performance in handling data with the concept drift property. Synthetic datasets are generated to provide a measurable reference.

\subsubsection{Synthetic Data Generation}
The ground truth is assume to have linear relationship with the features. 
For the $i$-th observation, let $y_i$ denote target value and $\bx_i$ denote the feature vector associated with each target.
\begin{equation}\label{eq:dataY}
y_i = \bbeta_i \cdot \bx_i^T + c +e_i
\end{equation}
Equation \ref{eq:dataY} illustrates the relation between the target and the features. 
Here $\bbeta_i$ is a coefficient vector with each of its elements corresponding to the elements in $\bx_i$, $c$ is a constant, $e_i$ is a random noise $e_i\sim N(0,1)$. The feature vector $\bx_i$ is generated as follow:
\begin{equation}\label{eq:dataX}
\bx_i = \nu z_i \bo + \widetilde{\be_i}
\end{equation}
Where $z_i\sim N(0,1)$, $\bo$ is a vector of the same length as $\bx_i$ and all its entries equal to 1, $\nu$ is a parameter to control the correlation between features, and $\widetilde{\be_i}$ is a noise vector with its $j$-th element $\widetilde{e_{ij}}\sim N(0,1)$. In this setup, the correlation between any two features is $\nu/(1+\nu)$. In our experiment $\nu$ is set to 0.5.

Let $j$ denote the element index in $\bbeta_i$ and $k$ denote the number of non-zero coefficients. The indices of non-zero coefficients are shifted every $l$ observations. Let $b$ denote the non-zero signal value. Then $\bbeta_i$ is constructed as:
\vspace{3mm}
\begin{equation}
\beta_{ij} =
    \begin{cases}
      b & \text{if $j \in (\floor*{\frac{i}{l}}:\floor*{\frac{i}{l}}+k)$}\\
      0 & \text{otherwise}
    \end{cases}    
\end{equation}
\begin{figure}[t]
\centering
\hspace{-4mm}
\includegraphics[width=0.45\linewidth]{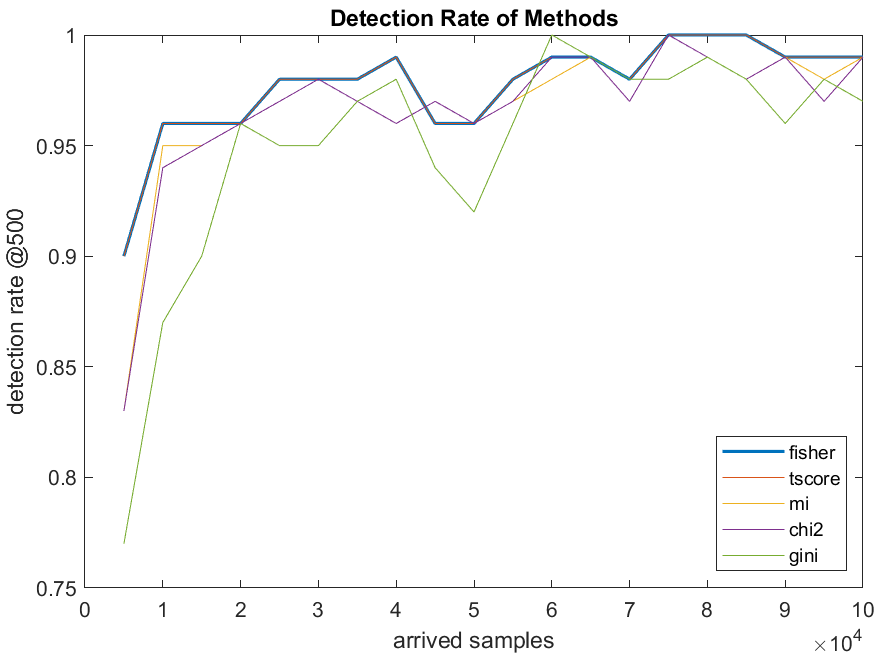}
\includegraphics[width=0.45\linewidth]{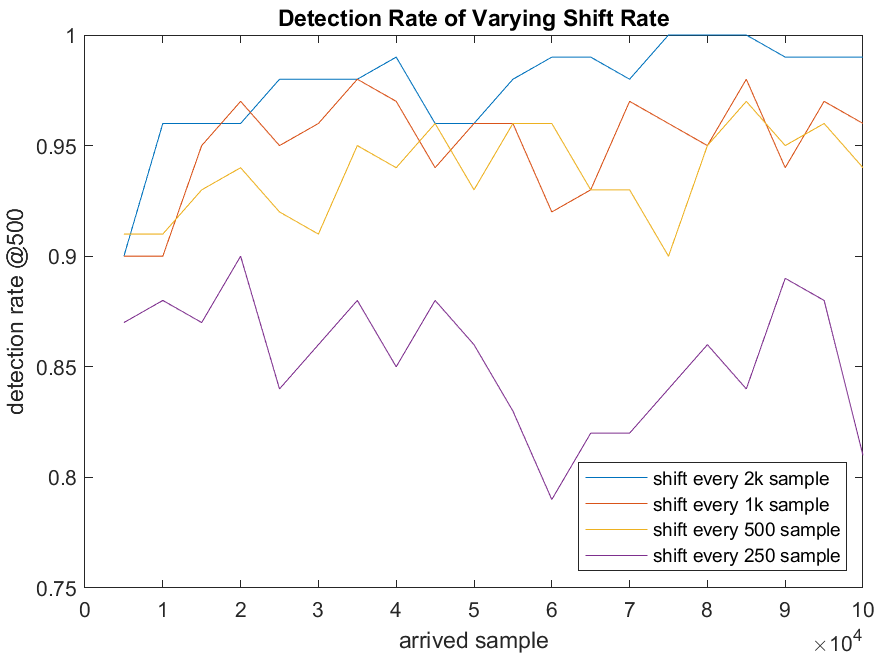}\\
\hspace{-4mm}\includegraphics[width=0.45\linewidth]{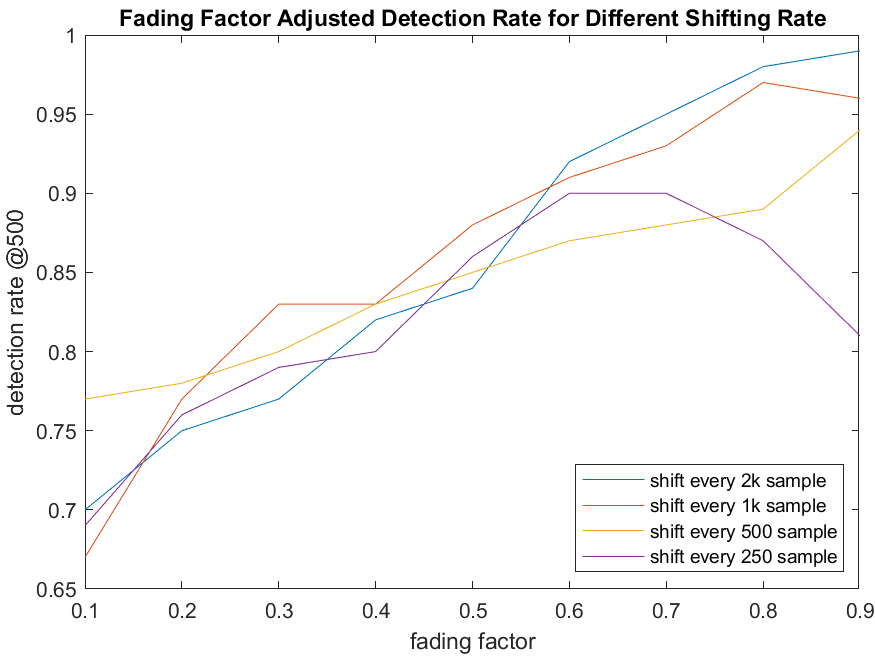}
\includegraphics[width=0.45\linewidth]{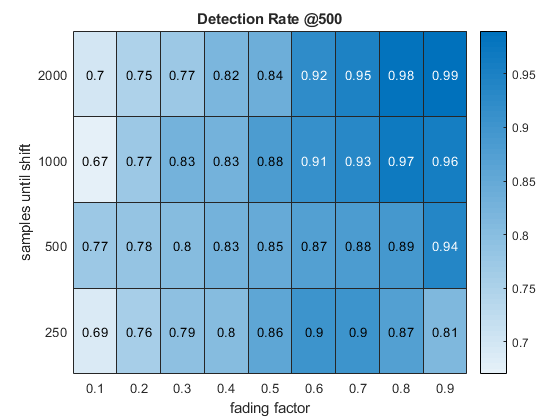}
\vskip -3mm
\caption{Top-left: detection rate by methods. 
Top-right: detection rate vs shifting rate.
Bottom-left: detection rate  by shifting rate adjusted by fading factor. 
Bottom-right: detection rate by shifting rate vs. fading factor.}\label{fig:synthetic}
\vspace{-6mm}
\end{figure}

In our experiments, we choose feature space to be 1000D, thus $\bx_i\in \RR^{1000}$. 
The number of true features $k$ is set to 100. The coefficient signal value is set to $b=1$. 
A total of 100,000 samples are generated. 
We change the value of $l$ (number of samples until shift) to control the concept drift level.

\subsubsection{Results}

In this section, we use detection rate @$k$ to measure how well the algorithm performed. Given a set of $k$ selected feature indices $FS$ and a set of true feature indices $TU$, the detection rate @$k$ is defined as.
\begin{equation}
DetRate @k = \frac{|FS\cap TU|}{|TU|}
\end{equation}
All $k$-s in this section are set to 500.

\begin{figure}[t]
\centering
\hspace{-4mm}
\includegraphics[width=0.45\linewidth]{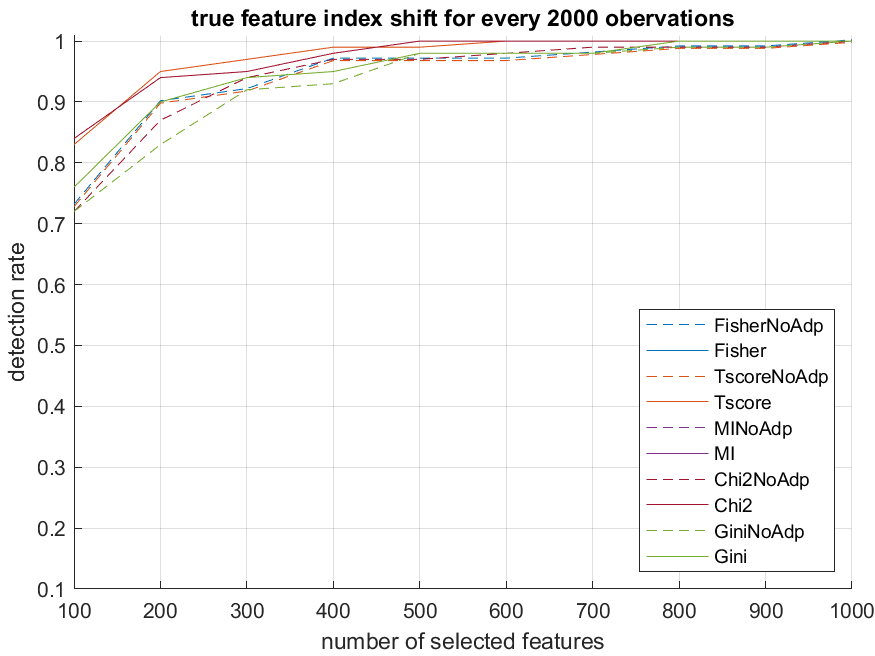}
\includegraphics[width=0.45\linewidth]{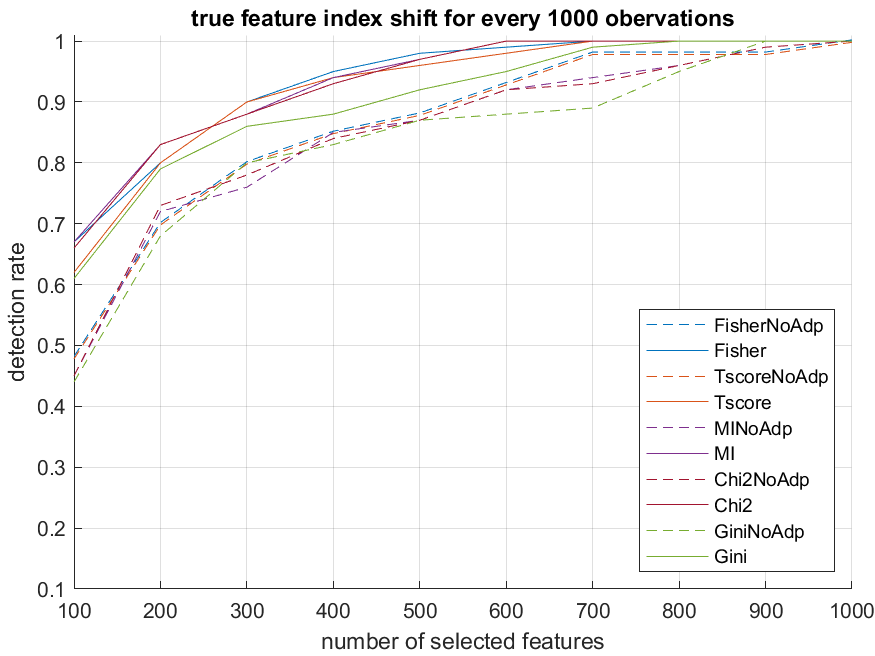}\\
\hspace{-4mm}\includegraphics[width=0.45\linewidth]{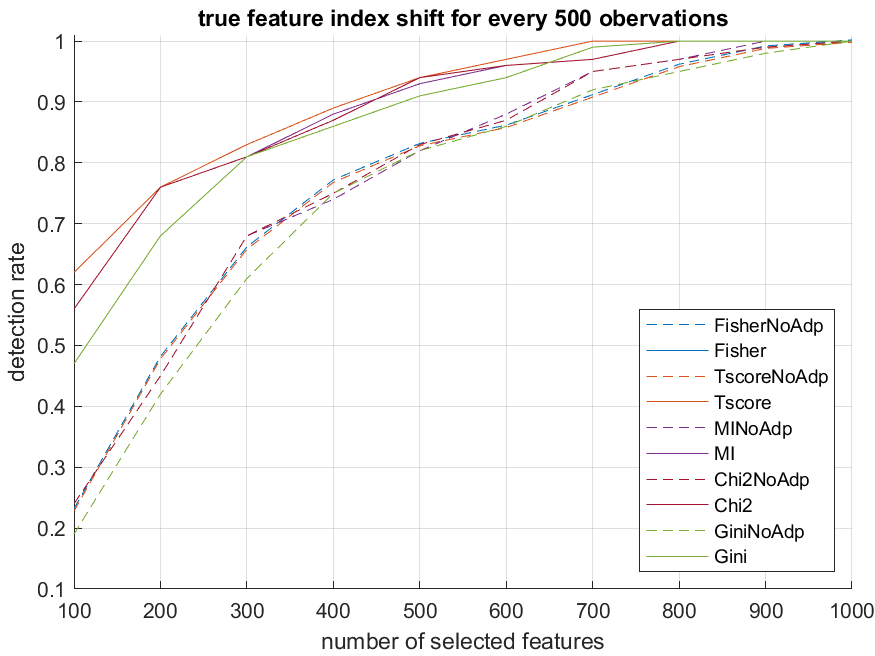}
\includegraphics[width=0.45\linewidth]{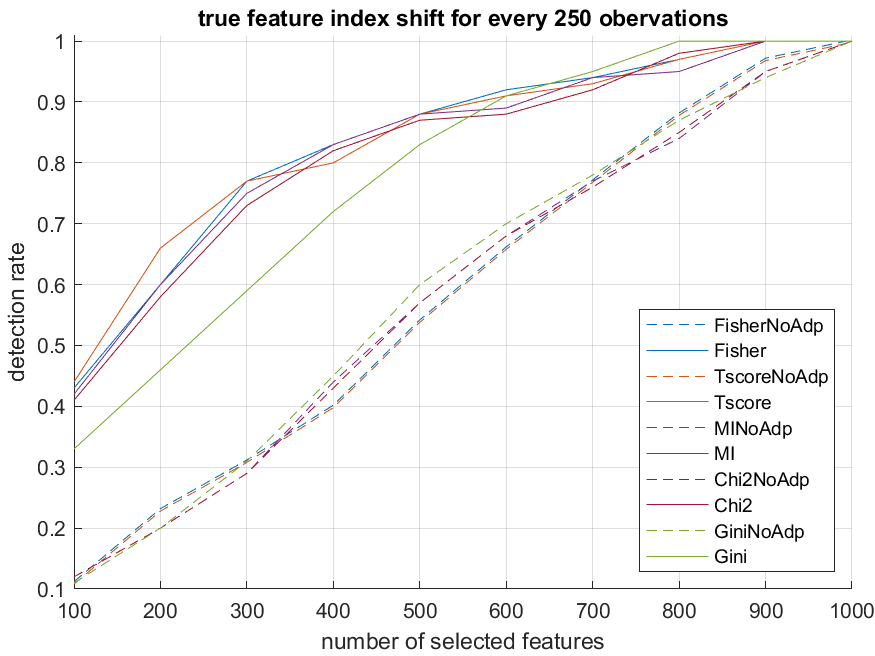}
\vskip -3mm
\caption{Influence of model adaptation on true variable detection rates for different rates of concept drift. Solid curves denote methods with model adaptation, dashed curves are methods without model adaptation.}\label{fig:syntheticInterval}
\vspace{-6mm}
\end{figure}
\paragraph{Performance Plots}
Fig \ref{fig:synthetic} visualizes the performance of online screening methods handling data with concept drift property. 
Top-left graph shows the overall performance of different screening methods. 
The shifting rate $l$ is fixed to 2000, and the fading factor $\alpha$ is fixed to $\alpha=0.9$. 
The five screening methods show very similar performance. 
However compared to the others, the Gini Index requires more samples to establish a good performance at the beginning.
The graph on the top-right shows how shifting rate affects the variable detection rate. 
The fading factor $\alpha$ is still fixed to $\alpha=0.9$. It is obvious that a fast shifting rate has a negative impact on the performance of the algorithm. 
On the bottom-left graph, it can be observed that by adjusting the fading factor $\alpha$, the detection rate can be improved even for data with a faster shifting rate.
The heatmap on the bottom-right shows the impact of fading factor on detection rate in detail.

In Fig \ref{fig:syntheticInterval}, it is shown in detail how the detection rate changes when different number of features are selected by online screening methods, as well as whether model adaptation helps improve the performance of the screening methods. 
Across all four shift rates that are tested, the screening methods with model adaptation outperform those without model adaptation. 
Generally, the slower the shift the fewer features need to be selected to provide a full detection.
Even in the fastest shift rate that is tested (shift every 250 samples), the screening methods with model adaptation still manage to detect all true features with fewer selected features and consistently detect more true features than the methods without model adaptation.

\subsection{Realistic Performance of Online Screening Methods}\label{sec:real}

This section provides an insight of the performance of online screening methods on real world large datasets. 
Different learners and screening methods are combined to generate prediction. 
The misclassification error rate is used in this section to measure the performance of the algorithm. 
For learners, Sparse FSA and SGD (stochastic gradient decent with log loss) are chosen.

\begin{table}[htb]
\vskip -1mm
\begin{center}
\caption{The datasets for realistic evaluation}\label{tab:dataset3}
\begin{tabular}{|l|c|c|c|c|}
\hline
Dataset &\thead{Learning type} &\thead{Feature type} &\thead{Number of\\ features} &\thead{Number of\\ observations}\\[0.5ex]
\hline
\hline
\href{https://www.jmlr.org/papers/v6/keerthi05a.html}{20NewsGroups} \cite{keerthi2005modified} &Classification &Binary &723,066 &11,862 \\[0.5ex]
\hline
Url \cite{ma2009identifying}&Classification &Continuous, Binary &3.2million &2million\\[0.5ex]
\hline
\end{tabular}
\end{center}
\vskip -6mm
\end{table}
\subsubsection{Data Sets}
Among the datasets showed in Table \ref{tab:dataset3}, Url data already be introduced in Section \ref{sec:onoff}. Here Url data from day 0 to 99 is used to train the model. 
The 20NewsGroups is an email content data that is generated according to \cite{keerthi2005modified}. The data originates from the UCI repository \cite{Dua:2019}. We also extracted the timestamp for each email and sorted the data in time order.

\subsubsection{Results}
For all tables in this section, the cell value ''random select'' indicates that a specified number of features are randomly selected as input for the learner. 
The first two rows in each table show the computing time for online screening methods. The ''Run Time'' column contains the training time only.

In Table \ref{tab:news20} is shown that for the 20NewsGroup data, some screening methods such as Mutual Information provide improvement to Sparse FSA. 
Moreover, in most cases, the screening methods with model adaptation give better performance than without model adaptation. 
For SGD, there is slight improvement given by bin-count based screening methods.

In Table \ref{tab:url}, it is shown that there is no improvement by applying screening methods on Sparse FSA. 
The training time is reduced due to smaller feature space caused by screening methods. 
When combined with SGD, the mean-variance based screening methods with model adaptation provide significant improvement on performance. 

\begin{table}[htb]
\vspace {-1mm}
\begin{center}
\caption{20 News Groups}
\label{tab:news20}
\begin{tabular}{|c|c|c|c|c|c|c|}\hline%
&&Selected & &Learner &Run Time &\\
Adaptation&Screening Method&Features  &Learner & Selected &Seconds &Error Rate\\ 
\hline
\hline
\csvreader[late after line=\\\hline]%
{20news.csv}{}
{\csvcoli & \csvcolii & \csvcoliii & \csvcoliv & \csvcolv& \csvcolvi& \csvcolvii}%
\end{tabular}
\end{center}
\vspace{-5mm}
\end{table}

\begin{table}[htb]
\vspace {-1mm}
\begin{center}
\caption{URL data}
\label{tab:url}
\begin{tabular}{|c|c|c|c|c|c|c|}\hline%
&&Selected & &Learner &Run Time &\\
Adaptation&Screening Method&Features  &Learner & Selected &Seconds &Error Rate\\ 
\hline
\hline
\csvreader[late after line=\\\hline]%
{URl.csv}{}
{\csvcoli & \csvcolii & \csvcoliii & \csvcoliv & \csvcolv& \csvcolvi& \csvcolvii}%
\end{tabular}
\end{center}
\vspace{-5mm}
\end{table}


\section{Conclusion}
In our studies of online screening methods, the moving average based online screening methods can be proved to have the same performance as their offline version and have the advantage of a faster speed and lower storage requirement. The experiments in Section \ref{sec:onoff} show that the bin-count based online screening methods can also achieve the same results as their offline version given the right $\epsilon$. Moreover, they can obtain faster or about the same computation speed compared to their offline versions with the right combination of $\epsilon$ and minibatch size. 

The results in Section \ref{sec:syn} give empirical evidences that adding model adaptation utility to screening methods can help improve their performance when data have concept drift property. It is shown that to some degree, adjusting fading factor can assist screening method to tackle datasets with high concept drift rate. 
The real data analysis in Section \ref{sec:real} further demonstrates the capability of online screening methods in dealing with real life large datasets with sparsity and possible concept drift.

Our results show that online screening methods with model adaptation is computationally efficient. They are useful in improving the model performance of complex data learning, especially when sample size and feature space are extremely large.

\section{Future Works}
Some criteria used by screening methods such as mutual information and Gini index are also applied in learning algorithms such as decision tree as a component to judge impurity. 
It triggers our interests to study whether online quantile based methods can improve the speed or even performance of decision trees. 

Throughout the years some works were done to describe the reconstruction and transformation from decision trees to neural network. 
Several mapping methods were mentioned in \cite{sethi1990entropy} \cite{sethi1991decision} \cite{kontschieder2015deep} and \cite{ioannou2016decision}. 
A new framework Neural Rule Ensembles (NRE) introduced in recent literature \cite{dawer2020neural} also focused on such a mapping strategy. 
It shows that any decision tree can be mapped into a set of neural rules, and the ensemble of neural rules can subsequently be trained using back-propagation. 
Therefore the expected improvement of training speed on decision trees can hopefully also reflect on neural network training.

\bibliographystyle{plain}
\bibliography{references}

\end{document}